\documentclass[10pt,journal]{IEEEtran}
\ifCLASSOPTIONcompsoc
\else
\fi
\ifCLASSINFOpdf
\else
\fi
\usepackage{amsmath}
\usepackage{array}
\usepackage{longtable}
\usepackage{amsmath}
\usepackage{epsfig,subfigure}
\usepackage{graphicx}

\usepackage{cite}
\usepackage{array}
\usepackage{algorithmic}
\usepackage{algorithm}
\usepackage{cite}
\usepackage{longtable}
\usepackage{setspace}

\usepackage{amsmath,amssymb,latexsym,url}

\begin{document}
%
\title{Contraction Principle based Robust Iterative Algorithms for Machine Learning}
\author{Rangeet Mitra, Amit Kumar Mishra,\textit{Senior Member, IEEE}}
\IEEEcompsoctitleabstractindextext{%
\begin{abstract}
\textbf{I}terative algorithms are ubiquitous in the field of data mining. Widely known examples of such algorithms are the least mean square algorithm, backpropagation algorithm of neural networks etc. 
  Our contribution in this paper is an improvement upon these iterative algorithms in terms of their respective performance metrics and robustness.
This improvement is achieved by a new scaling factor (motivated from contraction principle in topology) which is multiplied to the error term.
Our analysis also shows that, in essence, we are minimizing the corresponding LASSO (least absolute shrinkage and selection operator) cost function, which is the reason of its increased robustness. We also give closed from expressions for the number of iterations required for convergence and the MSE floor of the original cost function for a given minimum targeted value of the $L1$ norm. As a concluding theme based on stochastic subgradient algorithm, we give a comparison between the well known Dantzig selector and our algorithm based on contraction principle.
 By these simulations we attempt to show the optimality of our approach for any  widely used parent iterative optimization problem.
\end{abstract}

}

\maketitle

\IEEEdisplaynotcompsoctitleabstractindextext

%
\IEEEpeerreviewmaketitle

\section{Introduction}
In most engineering applications, we need to solve a system of linear equations. One of the popular approaches to solve a system of linear equations is based on least-squares.  An attractive way of solving system of equations is by iterative algorithms \cite{Rocky} due to computational simplicity and robustness of the solution (as compared to their batch counterparts).

Actually, iterative algorithms are ubiquitous in most of the machine learning algorithms. In this paper we pick up certain widely used iterative algorithms for our study: 1) Backpropagation Algorithm of neural Networks, 2)Least Mean Square Algorithm, 3)Kernel Least Mean Square Algorithm and 4)Dantzig Selector. One of the many ways of non-linear parameter estimation is Neural Networks (NN) \cite{Bishop,Haykos}. NN structure can consist of just one neuron, in  which case it is nothing but the widely known least mean square algorithm (not considering the activation function) \cite{Haykos}. In case there are many layers, the weights are adapted using the Backpropagation algorithm \cite{Haykos}. Apart from neural networks, there is another way of ``implicit" parameter estimation. These kernel parameter estimation techniques \cite{Bishop,Haykos,Liu2008} transform the data in a linearly non-separable indigenous space to infinite dimensional kernel space where it can be linearly separated. The beauty of kernel techniques is that we are saved from taking inner products in higher dimensional spaces by what we know as the ``kernel trick" \cite{Liu2008}, hence avoiding the curse of dimensionality. Primarily these kernel techniques were applied in nonlinear Support Vector Machines. Recently there has been much interest in ``kernelizing" other algorithms in adaptive filter theory. For example, the Recursive Least Squares (RLS) \cite{Sayed} now has a kernelized namesake called Kernel Recursive Least Squares (KRLS) \cite{KRLS}. Also, the well known Least Mean Square(LMS) \cite{Sayed} algortihm has an analog named the Kernel Least Mean Square(KLMS) \cite{Liu2008} algorithm. All the above are examples of iterative algorithms that find profound usage in everyday life. For example, one can hardly think of a channel-equaliser which does not use one of the above algorithms at its root \cite{Liu2008,Sayed}. Adaptive-denoising \cite{Sriram2005}, channel estimation \cite{Coleri2002} and adaptive beamforming in smart antennas \cite{Gaudes2007} are areas in which these iterative algorithms are used. 
Sometimes when the number of equations is less than the number of unknowns (i.e. the problem is ill-posed), we need to use what is called regularization \cite{Regularization}. Examples include Tikhonov regularization (for $L_{2}$-norm), and Least Absolute Shrinkage and Selection Operator (LASSO) for $L_{1}$ norm. The difference between the above two approaches lie in the assumption of regularization prior pdfs. For Tikhonov regularization the pdf is Gaussian while for LASSO the pdf is Laplacian.

One of the problems with these regularization techniques is increase in computational complexity of the optimization problem due to the regularization term. We tackle this problem by introducing a multiplication factor/variable step-size factor for the error noise motivated from a result in functional analysis. This achieves the same effect as LASSO/ridge regression. This factor would result in a lower cost function(both in the $L1$ and $L2$ norm sense) at convergence as compared to parent iterative algorithm without the factor.

 Rest of the paper is organised as follows.
Section II gives a review of the LMS algorithm, the KLMS algorithm and the Backpropagation algorithm in a unified way. Also, it gives details about a study theme which involves minimizing LASSO and Dantzig selector problems via subgradients. Section III describes our approach  briefly and gives the corresponding contraction principle based counterparts of the algorithms. Section IV gives results which validate our claims. Section V concludes the paper.

 \section{Examples of Some Common Iterative Algorithms in Machine Learning and the LASSO}
Iterative parameter estimation involves a parameter, say $\textbf{w}$
(please observe that we define it in an abstract manner. It can be a vector or a matrix depending on the context)
, which is estimated by the following rule:
\begin{equation}\label{adapt}
  \textbf{w}(n+1) = \textbf{w}(n) + \delta(n)
\end{equation}
In Eqn. \ref{adapt}, $n$ is the iteration number and $\delta$ is the adaptation noise.
Let us consider a data set $\{ \textbf{X}_{i}\}_{i=1}^{N}$ where $N$ is the data set size. The corresponding labels are $\{t_{i}\}_{i=1}^{N}$. Now, we present our three objects of study in this paper.

\subsubsection{The LMS Algorithm \cite{Sayed}}
In this case,
\begin{equation}\label{LMS}
  \delta(n) = \mu (t_{n} - \textbf{w}(n)^{T}\textbf{X}_{n})\textbf{X}_{n}
\end{equation}
Here $\mu$ denotes the step size. This is what is commonly known as the Widrow-Hopf adaptation rule.
\subsubsection{The KLMS Algorithm}
Invoking Eqn. \ref{adapt} in a different spirit, we may write it as follows,
\begin{equation}
 \textbf{w}(n+1) = \textbf{w}(n-1) + \delta(n) + \delta(n-1)
\end{equation}

Ignoring initial conditions, further decomposition would yield,

\begin{equation}
 \textbf{w}(n+1) =  \Sigma_{i=1}^{n} \delta(i)
\end{equation}

Hence,
\begin{equation}
 \langle\textbf{w}(n+1),\textbf{X}_{n}\rangle =  \Sigma_{i=1}^{n} \langle \delta(i), \textbf{X}_{n} \rangle
\end{equation}

This algorithm in \cite{Liu2008} invokes the kernel trick. These algorithms are meant for particularly linearly non-separable datasets. If we take the kernel inner product of the observation $\textbf{X}_{n}$ with the hypothesis $\textbf{w}$, we get the output $y_{n}$. If this inner product is a kernel inner product \cite{Liu2008}, we get the following equation from Eqn. \ref{LMS},

\begin{equation}\label{KLMS1}
  y(n+1) = \mu\Sigma_{i=0}^{n-1}e(i)\langle \textbf{X}_{i},\textbf{X}_{n}\rangle_{K}
\end{equation}

Here, we have the following expression for the error term
\begin{equation}
e(i) = t_{i} - y(i) , i=1,2...N.
\end{equation}

Eqn. \ref{KLMS1} can further be written in a recursive manner as follows,

\begin{equation}\label{KLMS2}
  y(n+1) = y(n) + \mu e(n)\langle\textbf{X}_{n-1},\textbf{X}_{n}\rangle_{K}
\end{equation}

Here the kernel inner product operator is defined as follows: given two matrices $\textbf{A} \varepsilon \mathbb{R}^{m\times n}$ and $\textbf{B} \varepsilon \mathbb{R}^{n\times p}$, the element belonging to the $i^{th}$ row and $j^{th}$ coloumn of the resultant matrix
is as follows,
\begin{equation}
C_{ij} = exp(-\frac{\|\textbf{A}_{i} - \textbf{B}_{j}\|^2}{\sigma})
\end{equation}

where $\textbf{C} \varepsilon  \mathbb{R}^{m\times p}= \langle \textbf{A} , \textbf{B}  \rangle_{K} .$ This $\sigma$ is a free parameter and its values can be found analytically, by cross validation and is problem dependent or more precisely dependent on the spread of the data.

This makes $\delta$ for this case to be,
\begin{equation}\label{KLMS3}
  \delta  =  \mu e(i)\langle\textbf{X}_{n-1},\textbf{X}_{n}\rangle_{K}
\end{equation}

\subsubsection{The Backpropagation Algorithm \cite{Bishop}}
In this case (thanks to the terminology, we had kept the meanings of eqn. \ref{adapt} abstract so that everything may fit in), $\{\textbf{w}\}$ is a set of neurons ordered by another index $j$. We consider the $\textbf{w}$ cascaded in any combination and $j$ denotes the layer number which goes from $1$ till $N$. Here $j=N$ is the final layer without loss of generality.

Neural Network training consists of two passes. In the first pass, the outputs are calculated. After that, the $\delta_{j}$ are estimated by a recursive algorithm called the back-propagation algorithm.

The technique of estimation of $\delta$ is given in \cite{Backprop}. We repeat the derivation given there for ease of the reader.

\begin{itemize}
\item Forward Pass - Calculate outputs for all neurons by sending the data through the network. Let the activations/ outputs be indexed by $a_{j}$.
\item Output Node - Update $\delta_{N}$ by the gradient of the cost function with respect to the output weights, i.e.,
 $\textbf{w}(N)^{n+1} = \textbf{w}(N)^{n} + \nabla_{\textbf{w}(N)} J$.
 \item Backpropagation - $\delta_{j} = \delta_{j+1}^{T}\textbf{w}^{j-1} \bullet \nabla_{\textbf{w}_{j}} f_{act}$. Here $\bullet$ is the Hadamard product and $f_{act}$ is an activation function.
\item Concurrently update $\textbf{w}(j)^{n+1} = \textbf{w}(j)^{n} + \delta_{j+1} \otimes a_{j}$.
\end{itemize}

\subsubsection{LASSO and Dantzig Selector}
LASSO \cite{Tibshirani2011} is a popular example of regularization, in which the cost function to be optimized is of the form,

\begin{equation}\label{lass}
J_{lasso} = \|\textbf{t} - \textbf{X} \textbf{w} \|_{2} + \lambda \|\textbf{w}\|_{1}.
\end{equation}

 $\lambda$ in the above equation is a regularization parameter. Also $\textbf{w}$ is the parameter to be estimated. $\textbf{t}$ and $\textbf{X}$ are the targets and  the data values respectively. One of the elegant ways to implement this algorithm is by using Interior point methods \cite{Boyd2004} and subgradients \cite{Rocky}.

Similarly the cost function for Dantzig selector is given by,
\begin{equation}\label{dant}
J_{Dantzig} = \|\textbf{t} - \textbf{X} \textbf{w} \|_{\infty} + \lambda \|\textbf{w}\|_{1}.
\end{equation}

The directions for descent for these non-smooth cost functions are given by their subgradients \cite{Rocky} as follows,
\begin{equation}
\nabla J_{lasso} = (\textbf{t}_{n} - \textbf{X}_{n}^{T} \textbf{w})\textbf{X}_{n} + \lambda \tt{sign}(\textbf{w})
\end{equation}
and,
\begin{equation}\label{grad_dant}
\nabla J_{Dantzig} =  (\textbf{t}_{n}^{max} - (\textbf{X}_{n}^{T} \textbf{w})^{max})\textbf{X}_{n}^{max} + \lambda \tt{sign}(\textbf{w})
\end{equation}
A noteworthy comment is as follows: as the subgradient of the $L_{\infty}$ norm of the error is the convex hull of the individual gradients, we may choose any one of them as a valid subgradient. But instead of choosing randomly, we may choose according to some criterion like the direction with maximum cost (and hence needs to be penalized) etc as indicated as subscript $max$ in eq. \ref{grad_dant}.

Finally, we would iterate by,
\begin{equation}\label{subgrad1}
  \textbf{w}(n+1) = \textbf{w}(n) + \nabla J_{lasso}
\end{equation}
or,
\begin{equation}\label{subgrad1}
  \textbf{w}(n+1) = \textbf{w}(n) + \nabla J_{Dantzig}
\end{equation}

\subsubsection{Recursive Least Squares (RLS)}
The steps of the RLS recursion given in \cite{Sayed} for the given forgetting factor $\lambda$ are enumerated as follows,

\begin{algorithm}                      
\caption{Parameter Estimation for  $m$ for $\{\textbf{x}\}_{i=1}^{N}$}          
\label{basic}                           
\begin{algorithmic}[1]
\FOR{$n = 1$ to $N$}
\STATE $y(n) = \textbf{w}^{T}\textbf{x}$. $e(n) = (t(n)-y(n))$.
\STATE $\textbf{k}(n) = \textbf{P}(n-1) \textbf{x}/(\lambda + \textbf{x}^{T}\textbf{P}(n-1)\textbf{x})$
\STATE $\textbf{w}(n) = \textbf{w}(n-1) + \textbf{k}(n)e(n)$
\STATE $\textbf{P}(n)=\frac{1}{\lambda}(\textbf{P}(n-1) - \textbf{k(n)}\textbf{x}^{'}\textbf{P}(n-1))$
\ENDFOR
\end{algorithmic}
\end{algorithm}

Here $\textbf{P}(n)$ is an estimate of the inverse of the autocorrelation matrix at time $n$.

\section{Proposed Algorithm}
According to the contraction mapping theorem presented in \cite{Gamelin}, if $T$ is a contraction on a Banach Space, and we are dealing with a recursion of the form,

\begin{equation}\label{cprip}
  x_{n+1} = T(x_{n}) + u
\end{equation}

then,

\begin{equation}\label{cprip1}
  \|x-x^*\| \leq\ \frac{|T\|^{m}\|u\|}{1-\|T\|}
\end{equation}

where, $x^*$ is the fixed point of the iteration.

In our case $u\rightarrow0$ and,
\begin{equation}\label{cprip2}
  T(\textbf{w}(n)) = \textbf{w}(n) + \delta
\end{equation}

This is true as with every iteration, the weights move closer and closer towards its equilibrium point(s) (which hopefully is unique depending on the algorithm, or all equilibrium points are almost equally preferable to us).

We assume infinitesimal adaptation noise which is due to the step size which is generally chosen to be small. This assumption is sometimes invoked to analyze iterative algorithms as in \cite{Sayed}. With this assumption,
\begin{equation}\label{norm_T}
 Lt_{\delta\rightarrow0} \|T\| = Lt_{\delta\rightarrow0}\|\textbf{w}(n) + \delta\| = \|\textbf{w}(n)\|
\end{equation}

Hence the residual may be attributed to $\textbf{u}$. Hence by the tight inequality $\|\textbf{w}(n) + \delta\| \leq \parallel\textbf{w}(n)\parallel + \parallel\delta\parallel$, we can assign $\parallel\textbf{u}\parallel = \parallel\delta\parallel$.

Hence our result becomes for LMS and NN approaches,
\begin{equation}\label{cprip3}
  \|\textbf{w}-\textbf{w}^*\|_{1} \leq\ \frac{|\textbf{w}\|_{1}^{m}\|u\|_{1}}{1-\|\textbf{w}\|_{1}}
\end{equation}
For the KLMS approach, the result is,
\begin{equation}\label{cprip4}
  \|y(n)-y(n)^*\|_{1} \leq\ \frac{|y(n)|^{m}\|u\|_{1}}{1-|y(n)|}
\end{equation}

We can observe the tendency for the deviation to increase as $\|\textbf{w}\|_{1}$ goes near $1$. Also, there is a tendency to diverge if $\|\textbf{w}\|_{1}>1$. Hence our proposal is to use a correcting factor multiplied to the error term.

This correcting factor has the unwanted $(1-\parallel\textbf{w}\parallel_{1})$ in the numerator and $\parallel\textbf{w}\parallel_{1}$ in the denominator. This gives a \textbf{modified Widrow-Hoff paradigm} as follows.

\begin{equation}\label{adapt_mod}
  \textbf{w}(n+1) = \textbf{w}(n) + (1+\frac{(1-\parallel\textbf{w}\parallel_{1})}{\parallel\textbf{w}\parallel_{1}})\delta(n)
\end{equation}

In Eqn. \ref{adapt_mod}, the $\textbf{w}(n)$ refers to the weights of the hyperplane at time instant $n$.

Similarly, a \textbf{modified Backpropagation Algorithm} would be given by the following equation.

\begin{equation}\label{adapt_modq}
  \textbf{w}(j) = \textbf{w}(j) + (1 + \frac{(1-\parallel\textbf{w}(j)\parallel_{1})}{\parallel\textbf{w}(j)\parallel_{1}})\delta_{j}(n)
\end{equation}

In the above Eqn. \ref{adapt_modq}, $\textbf{w}(j)$ denotes the weight matrix of the $j^{th}$ layer and the $\delta_{j}(n)$ is found by the backpropagation algorithm in \cite{Backprop}.

Also, a \textbf{modified KLMS} algorithm would be given by,
\begin{equation}\label{adapt_mody}
  y(n+1) = y(n) + (1 + \frac{1 - |y(n)|}{|y(n)|}) e(n)<\textbf{X}_{n-1},\textbf{X}_{n}>_{K}
\end{equation}
Contraction principle based LASSO and Dantzig selector problems may be handled by the following equations.
\begin{equation}\label{adapt_lasso}
  \textbf{w}(n+1) = \textbf{w}(n) + (1 + \frac{(1-\parallel\textbf{w}\parallel_{1})}{\parallel\textbf{w}\parallel_{1}})\nabla J_{lasso}
\end{equation}
\begin{equation}\label{adapt_dantzig}
  \textbf{w}(n+1) = \textbf{w}(n) + (1 + \frac{(1-\parallel\textbf{w}\parallel_{1})}{\parallel\textbf{w}\parallel_{1}})\nabla J_{Dantzig}
\end{equation}

Here, in Eqn. \ref{adapt_lasso} and \ref{adapt_dantzig} $\nabla$ denotes a subgradient of those functionals.

Similarly, the \textbf{modified RLS} will be given as follows,

\begin{algorithm}                      
\caption{Parameter Estimation for  $m$ for $\{\textbf{x}\}_{i=1}^{N}$}          
\label{basic}                           
\begin{algorithmic}[1]
\FOR{$n = 1$ to $N$}
\STATE $y(n) = \textbf{w}^{T}\textbf{x}$. $e(n) = (t(n)-y(n))$.
\STATE $\textbf{k}(n) = \textbf{P}(n-1) \textbf{x}/(\lambda + \textbf{x}^{T}\textbf{P}(n-1)\textbf{x})$
\STATE $\textbf{w}(n) = \textbf{w}(n-1) + (1 + \frac{1-\|\textbf{w}(n)\|_{1}}{\|\textbf{w}(n)\|_{1})}\textbf{k}(n)e(n)$
\STATE $\textbf{P}(n)=\frac{1}{\lambda}(\textbf{P}(n-1) - (1+\frac {1-\|(\textbf{P}(n-1)\|_{1}}{\|(\textbf{P}(n-1)\|_{1}})\textbf{k(n)}\textbf{x}^{'}\textbf{P}(n-1))$
\ENDFOR
\end{algorithmic}
\end{algorithm}

\subsection{Modeling the Problem in an Operator Theoretic Perspective}
In this section, we give step by step details of our problem formulation based on the Contraction principle \cite{Gamelin} for the ease of reader. Let us assume a dynamical system which evolves as follows,
\begin{equation}\label{base_1}
  \textbf{w}(n+1) = T(\textbf{w}(n)) + \delta
\end{equation}
Let us assume that $\textbf{w}^{*}$ is an equilibrium point. If that is the case, and considering $T(\textbf{w}^*) = \textbf{w}^*$ (due to assumption of it being an equilibrium point), we subtract $\textbf{w}^*$ from both sides of the equation.
\begin{equation}\label{base_2}
  \textbf{w}(n+1) - \textbf{w}^{*} = T(\textbf{w}(n)) - \textbf{w}^{*} + \delta
\end{equation}
or,
\begin{equation}\label{base_3}
  \textbf{w}(n+1) - \textbf{w}^{*} = T(\textbf{w}(n)) - T(\textbf{w}^{*}) + \delta
\end{equation}
We define a new variable $\tilde{\textbf{w}}(n) = \textbf{w}(n) - \textbf{w}^{*}$. Hence Eqn. \ref{base_3} becomes,
\begin{equation}\label{base_4}
  \tilde{\textbf{w}}(n+1) = T(\tilde{\textbf{w}}(n)) + \delta
\end{equation}
Thus,
\begin{equation}\label{base_5}
  \tilde{\textbf{w}}(1) = T(\tilde{\textbf{w}}(0)) + \delta
\end{equation}
\begin{equation}\label{base_6}
  \tilde{\textbf{w}}(2) = T^{2}(\tilde{\textbf{w}}(0)) + T \delta + \delta
\end{equation}
\begin{equation}\label{base_7}
  \tilde{\textbf{w}}(3) = T^{3}(\tilde{\textbf{w}}(0)) + T^2 \delta + T \delta +\delta
\end{equation}
and so on till,
\begin{equation}\label{base_8}
  \tilde{\textbf{w}}(n) = T^{n}(\tilde{\textbf{w}}(0)) + T^{n-1} \delta + . . . T \delta +\delta
\end{equation}
also,
\begin{equation}\label{base_9}
  \tilde{\textbf{w}}(n+k) = T^{n+k}(\tilde{\textbf{w}}(0)) + T^{n+k-1} \delta + . . . T \delta +\delta
\end{equation}
Let us assume of large enough $n$ the algorithm reaches a fixed point such that $T^{n+k}(\tilde{\textbf{w}}(0))
=T^{n}(\tilde{\textbf{w}}(0))$.
Hence,
\begin{equation}\label{base_10}
 \parallel\tilde{\textbf{w}}(n) - \tilde{\textbf{w}}(n+k)\parallel = \parallel T^{n+k-1} \delta + . . . T^{n} \delta \parallel
\end{equation}

This will be upper bounded by,

\begin{equation}\label{base_11}
 \parallel\tilde{\textbf{w}}(n) - \tilde{\textbf{w}}(n+k)\parallel \leq \frac{\parallel T \parallel^{n} \parallel \delta \parallel}{1 - \parallel T \parallel}
\end{equation}

Then by definition of $\tilde{\textbf{w}}(n)$,
\begin{equation}\label{base_12}
 \parallel\textbf{w}(n) - \textbf{w}(n+k)\parallel \leq \frac{\parallel T \parallel^{n} \parallel \delta \parallel}{1 - \parallel T \parallel}
\end{equation}

Assuming convergence at $(n+k)^{th}$ iteration we get Eqn. \ref{cprip1}.

\subsection{Dependence on the Indicator Variable - Modification by Normalization with Dual Norm of the Data}\label{normalization}

In a classification problem, generally we would desire,
\begin{equation}\label{base_13}
  \langle \textbf{w},\textbf{x}\rangle = \mp 1
\end{equation}
depending on which class they belong. Here the inner product can be a linear or kernel (in which case it is implicit).
Hence by Holder's inequality,
\begin{equation}\label{base_14}
  \|\langle \textbf{w},\textbf{x}\rangle \|_{1} \leq \|\textbf{w}\|_{1} \|\textbf{x}\|_{\infty}
\end{equation}
This gives,
\begin{equation}\label{base_15}
sup_{\|\textbf{x}\|_{\infty}=1}  \frac{\|\langle \textbf{w},\textbf{x}\rangle \|_{1}}{\|\textbf{x}\|_{\infty}} = \|\textbf{w}\|_{1}
\end{equation}
In the situation that $\|\textbf{x}\|_{\infty} \neq 1$, we must divide the value of $\parallel T \parallel$ by $\|\textbf{x}\|_{\infty}$ (to follow definition of a norm given in \cite{Gamelin}). In all previous sections, the derivations were assuming $\textbf{x}$ to be in the unit circle.

Hence our norm for the $\parallel T \parallel$ is,

\begin{equation}\label{base_16}
  \parallel T \parallel = \frac{\|\textbf{w}\|_{1}}{\|\textbf{x}\|_{\infty}}
\end{equation}
Equivalently, for the kernel case, our factor would be  $\frac{1-\frac{|y(n)|}{\|\textbf{x}\|_{\infty}}}{\frac{|y(n)|}{\|\textbf{x}\|_{\infty}}}$. Please note that we don't need inner products and Holder's inequality to justify our cause of normalization. Such arguments are valid only when $\textbf{w}$ is a vector. However, when $\textbf{w}$ is a matrix, we need the output of the operator to be bounded within the unit circle. Hence, the same normalization factor may be justified by the duality of the $L1$ and $L_{\inf}$ norms.

\subsection{Relationship Between Step-Size and Upper Bound on  L1 norm }
Let us assume that the $L1$ norm is a small number $\epsilon$ at convergence. The step-size is inversely proportional to what is called the ``time-constant" \cite{Sayed} (a measure of speed of convergence of the algorithm). Hence, if we want convergence in less than or equal to $N$ iterations and some desired $MSE$ floor, we give the following (conflicting) design equations,
\begin{equation}\label{des1}
  k1 \frac{\epsilon}{\mu(1-\epsilon)} \leq N
\end{equation}

\begin{equation}\label{des2}
  k2 \frac{\mu(1-\epsilon)}{\epsilon} = MSE
\end{equation}
Here $k1$ and $k2$ are given in \cite{Sayed}.

\subsection{Equivalence to LASSO}

From Eqn. \ref{cprip3},
\begin{equation}\label{cpripq}
  \|\textbf{w}-\textbf{w}^*\|_{1} \leq\ \frac{|\textbf{w}\|_{1}^{m}\|u\|_{1}}{1-\|\textbf{w}\|_{1}}
\end{equation}
By triangular inequality from linear algebra,
\begin{equation}\label{cpripl}
\|\textbf{w}-\textbf{w}^*\|_{1} \leq\  \|\textbf{w}\parallel_{1}+\parallel-\textbf{w}^*\|_{1} \leq\ \frac{|\textbf{w}\|_{1}^{m}\|u\|_{1}}{1-\|\textbf{w}\|_{1}}
\end{equation}
As $\textbf{w}^*$ is independent of \textbf{w} analytically, hence minimizing the upper bound would result in minimizing the upper bound for the $L_{1}$ norm of the weights. Hence our algorithm, achieves a huge role in minimizing the upper bound(i.e. the L1 ball) in which the weights lie.

\subsection{Useful Properties I- Concavity in \textbf{w} within the unit circle}

Let us assume two candidate weights $\textbf{w}_{0}$ and $\textbf{w}_{1}$.

From convexity of any norm,

\begin{equation}
\lambda\parallel\textbf{w}_{0}\parallel + (1-\lambda)\parallel\textbf{w}_{1}\parallel \geq \parallel\lambda\textbf{w}_{0} + (1-\lambda)\textbf{w}_{1}\parallel
\end{equation}

Hence as norm is positive definite (it is semi-definite but the weights of an iterative algorithm are seldom 0),
\begin{equation}
\frac{1}{\lambda\parallel\textbf{w}_{0}\parallel + (1-\lambda)\parallel\textbf{w}_{1}\parallel} \leq \frac{1}{\parallel\lambda\textbf{w}_{0} + (1-\lambda)\textbf{w}_{1}\parallel}
\end{equation}

Subtracting 1 from both sides,
\begin{equation}
\frac{1}{\lambda\parallel\textbf{w}_{0}\parallel + (1-\lambda)\parallel\textbf{w}_{1}\parallel}-1 \leq \frac{1}{\parallel\lambda\textbf{w}_{0} + (1-\lambda)\textbf{w}_{1}\parallel} -1
\end{equation}

Please note that if the parent adaptation is convex and stable, each iteration would be a contraction towards the optimal weight. Hence the norm of the (normalized) weights (see Section. \ref{normalization}) should be less than or equal to unity always. Thus this proposed factor is always positive definite in normal cases.


\subsection{Conclusion From the Above Analysis}
\begin{figure}
  \centering
  \includegraphics[scale=0.5]{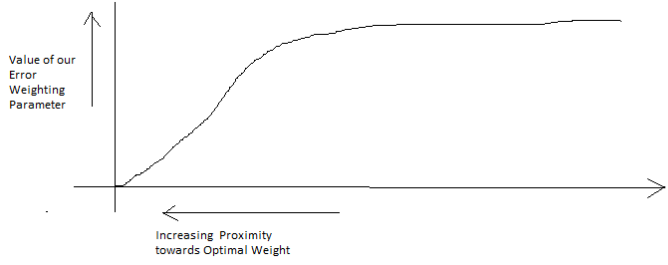}\\
  \caption{The Predicted Behavior of the Variable Step Size }\label{hypothesis}
\end{figure}

This shows that in effect, we get a curve similar to the one shown in Figure. \ref{hypothesis}. This does not come as a surprise; such curves have been reported in the literature \cite{Liyi2009}. However, the beauty of our approach is that our step size is
adjusted in such a manner that the LASSO cost function is minimized and hence has a nice ``regularization" aspect to it. These good properties also help us in selecting the most optimal equilibrium point (in a least-L1 norm sense) when there are many.

\section{Results}
\subsection{Experimental Setup}
In  Figure \ref{LMSS}, a BPSK constellation was generated and passed through a channel [-0.3,0.8]. After that $25dB$ Additive White Gaussian Noise (AWGN) was added. This data was input to the LMS and the modified LMS algorithm.

In Figure \ref{KLMSS}, the same BPSK constellation was passed through the same 2-tap channel $[-0.3, 0.8]$. After that it was passed through a non-linearity $g(x) = 1 - 0.9x^2$. After that $25 db$ noise was added.

In Figure \ref{NNN_chan2}, the 2-tap channel was the same one that was used in \cite{Liu2008}. The coefficients are $[1,0.5]$ and the $SNR$ used was $30dB$. The same non-linearity $g(x) = 1-0.9x.^2$ was applied.

For the figure in Figure. \ref{NNN}, the 2-tap channel was changed to $[1,1]$ and the non-linearity was changed to $h(x) = 1 - 0.5x^2$, just for the sake for variety. Here we increase the inter-symbol interference and reduce the non-linearity.

\subsection{Observations}
\begin{figure}
  \centering
  \includegraphics[scale=0.5]{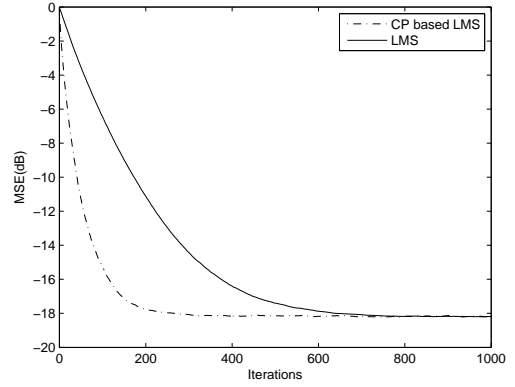}
  \caption{Performance Comparison with LMS}\label{LMSS}
\end{figure}

From Figure \ref{LMSS}, we observe that the proposed algorithm converges faster to the same testing MSE than the original LMS algorithm. This is a manifestation of the superior performance of the modified error term.

\begin{figure}
  \centering
  \includegraphics[scale=0.5]{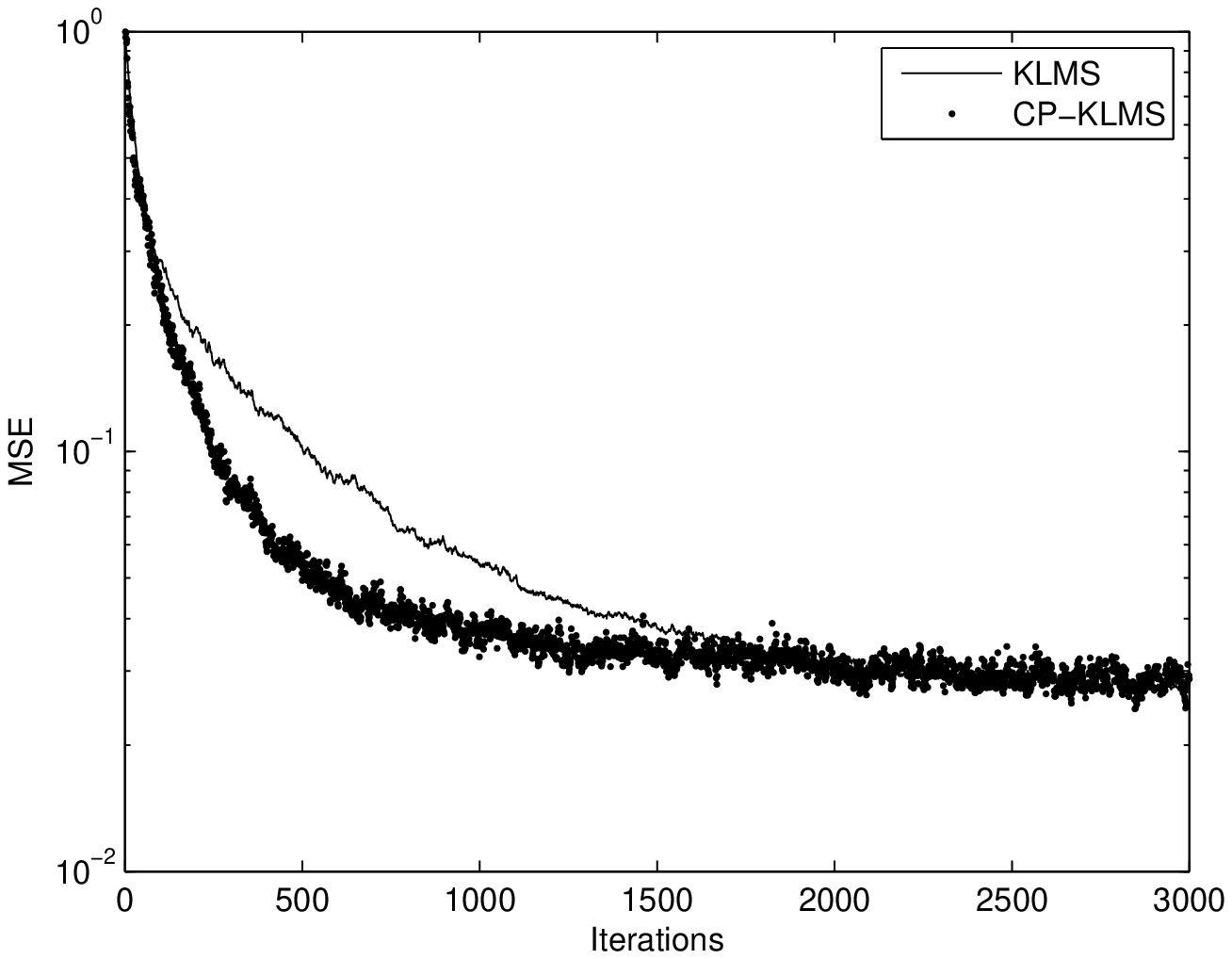}\\
  \caption{Performance Comparison with KLMS}\label{KLMSS}
\end{figure}

From Figure \ref{KLMSS}, we again see a significantly faster convergence rate. Also, the test error reaches a lower floor for our proposed algorithm than the original KLMS algorithm.

\begin{figure}
  \centering
  \includegraphics[scale=0.5]{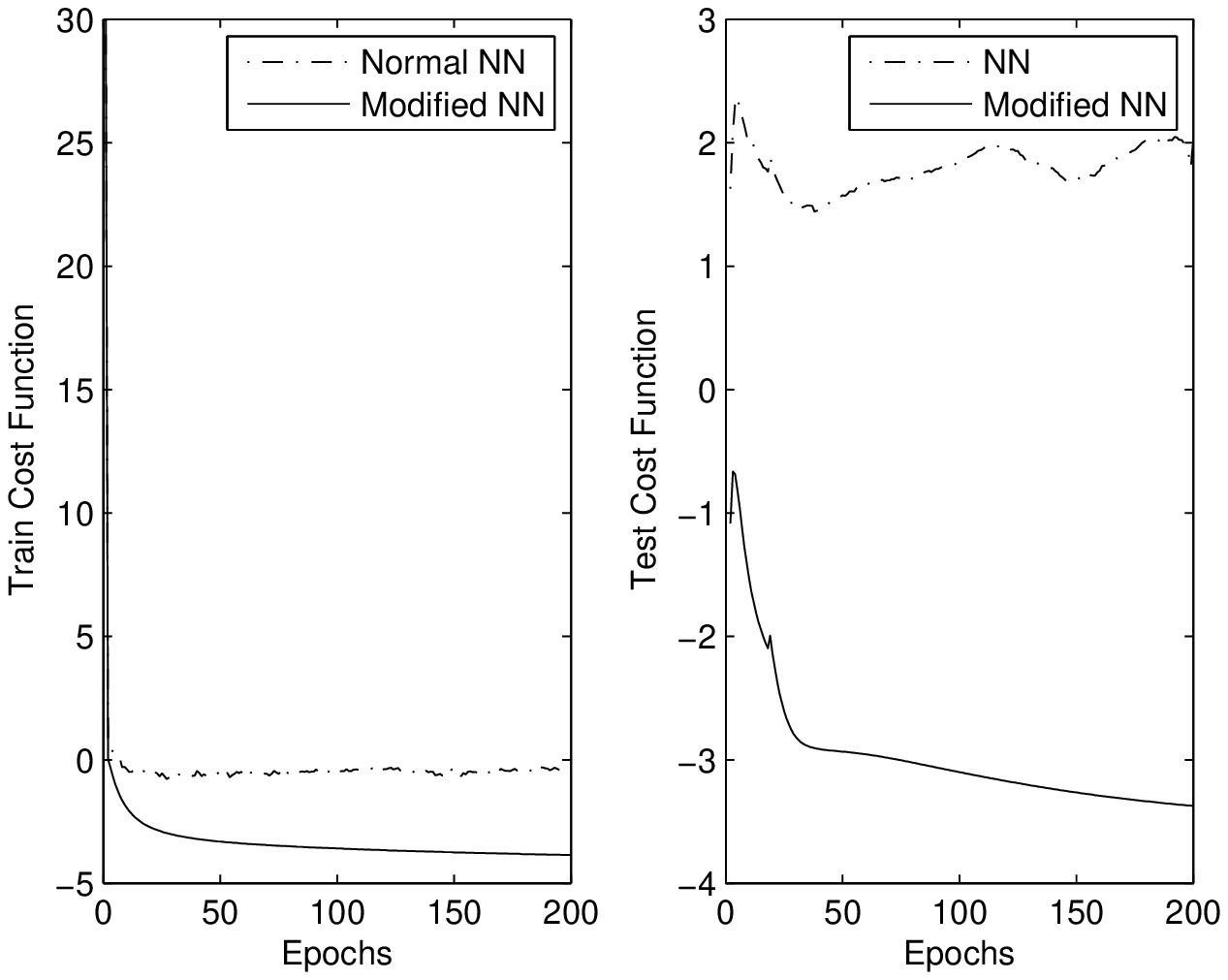}\\
  \caption{Performance Comparison with NN}\label{NNN}
\end{figure}

From Figure \ref{NNN}, we see that again the training and testing cost function values are much lower for our proposal. Also, we can see that our proposal is more robust. Also, the original Neural-Network configuration goes from one local minima in the testing cost-function to another during testing. However our scaling factor maintains its value in a steady manner. We understand that its value should increase after some epochs. But till 200 epochs we did not get any increase in the values which is an example of its robustness.

\begin{figure}
  \centering
  \includegraphics[scale=0.5]{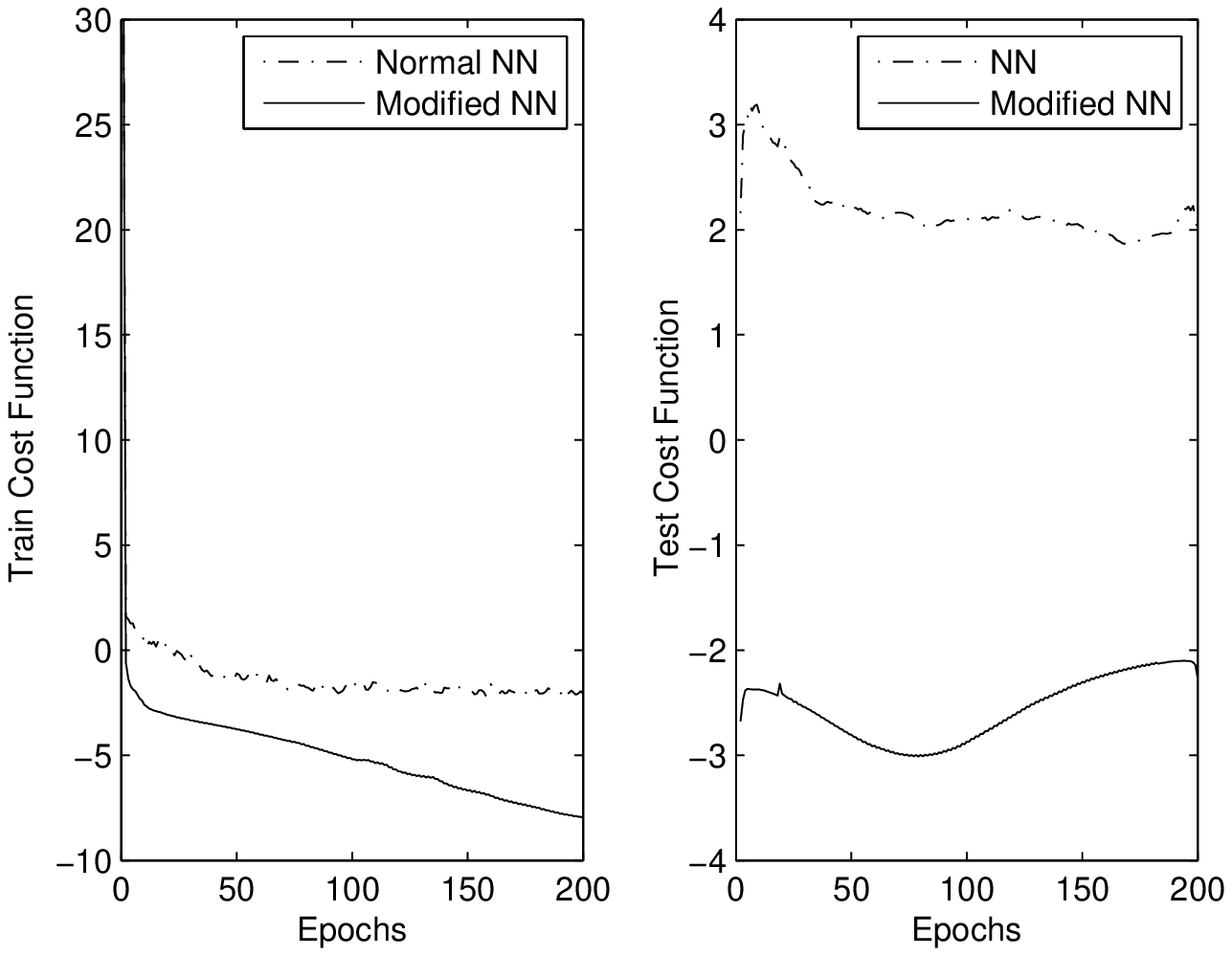}\\
  \caption{Performance Comparison with NN}\label{NNN_chan2}
\end{figure}

From Figure \ref{NNN_chan2}, we see that the training and testing cost function values are lower for our proposal. We see an increase in the testing MSE as a function of epochs in this case. However, it does not increase by more than $1dB$ in 150 epochs and stays well below the original NN curve.

\begin{figure}
  \centering
  \includegraphics[scale=0.5]{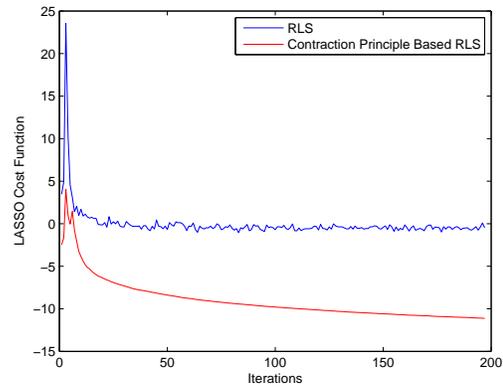}\\
  \caption{Performance Comparison between RLS and Modified-RLS}\label{dant_fig}
\end{figure}

From Figure \ref{dant_fig}, we find that the contraction principle based stochastic subgradient LASSO is converging faster than the original stochastic subgradient LASSO. We used synthetic data for this particular simulation by generating random matrices from a uniform distribution and fitting a regression between them.

\begin{figure}
  \centering
  \includegraphics[scale=0.5]{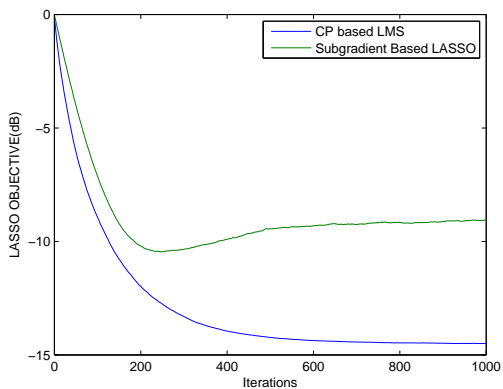}\\
  \caption{Performance Comparison with Dantzig Selector(method used:Stochastic Subgradient Algorithm)}\label{rls}
\end{figure}

From Figure \ref{rls}, we find that our contraction principle based RLS variant outdoes the conventional RLS algorithm. A Binary Phase Shift Keying(BPSK) constellation is randomly generated and passed through an FIR filter with coefficients $h=0.4,0.5,-0.4,1$.
Consequently, 20dB white Gaussian noise is added. With this dataset, we compared the conventional RLS and our RLS variant with the LASSO objective as the benchmark.

It is also worth mentioning that the above curves are not instantaneous curves; they have been obtained by averaging over at least 25 monte carlo simulations.

\section{Conclusion}
A number of common iterative algorithms have been evaluated and compared with their newly proposed contraction principle based variants. In all the scenarios we get a performance boost after applying our modification to the parent algorithms. Also, we show a relation between our approach and the LASSO. These results, which show a lower testing error on all occasions certify the superiority of our approach.

\appendices




%

\bibliography{paper}
\bibliographystyle{IEEEtran}

\end{document}